\documentclass[conference]{IEEEtran}
\IEEEoverridecommandlockouts

\usepackage{cite}
\usepackage{amsmath,amssymb,amsfonts}
\usepackage{graphicx}
\usepackage{textcomp}
\usepackage{xcolor}
\usepackage{soul}
\usepackage{cite}
\usepackage{balance}
\usepackage{bm}
\usepackage{algorithm} 
\usepackage{eucal}
\usepackage[algo2e]{algorithm2e} 
\usepackage{algpseudocode}
\usepackage{booktabs} 
\usepackage[usestackEOL]{stackengine}
\usepackage{url}
\usepackage{hyperref}
\usepackage{subfig}

\usepackage{array}
\usepackage{multirow}
    
\begin{document}


\title{Discern-XR: An Online Classifier for Metaverse Network Traffic
}


\author{\IEEEauthorblockN{Yoga Suhas Kuruba Manjunath\IEEEauthorrefmark{1},
Austin Wissborn\IEEEauthorrefmark{1}, Mathew Szymanowski \IEEEauthorrefmark{1}, \\Mushu Li \IEEEauthorrefmark{2}, Lian Zhao\IEEEauthorrefmark{1},  and Xiao-Ping Zhang \IEEEauthorrefmark{3}}
\IEEEauthorblockA{\IEEEauthorrefmark{1}Department of Electrical, Computer \& Biomedical Engineering, Toronto Metropolitan University, Toronto, Canada\\
\IEEEauthorrefmark{2}Department of Computer Science and Engineering, Lehigh University, Bethlehem, PA, USA\\
\IEEEauthorrefmark{3} Shenzhen Key Laboratory of Ubiquitous Data Enabling, Tsinghua Shenzhen International \\Graduate School, Tsinghua University, \\
Email: \{yoga.kuruba@torontomu.ca, austin.wissborn@torontomu.ca, mszymanowski@torontomu.ca,\\ mul224@lehigh.edu, l5zhao@torontomu.ca, xpzhang@ieee.org\}}}




\maketitle

\begin{abstract}
In this paper, we design an exclusive Metaverse network traffic classifier, named \textit{Discern-XR}, to help Internet service providers (ISP) and router manufacturers enhance the quality of Metaverse services. Leveraging segmented learning, the Frame Vector Representation (FVR) algorithm and Frame Identification Algorithm (FIA) are proposed to extract critical frame-related statistics from raw network data having only four application-level features. A novel Augmentation, Aggregation, and Retention Online Training (A2R-OT) algorithm is proposed to find an accurate classification model through online training methodology. In addition, we contribute to the real-world Metaverse dataset comprising virtual reality (VR) games, VR video, VR chat, augmented reality (AR), and mixed reality (MR) traffic, providing a comprehensive benchmark. Discern-XR outperforms state-of-the-art classifiers by 7\% while improving training efficiency and reducing false-negative rates. Our work advances Metaverse network traffic classification by standing as the state-of-the-art solution.

\end{abstract}


\section{Introduction}

The Metaverse, a concept combining virtually shared spaces and extended reality (XR), encompasses VR, AR, and MR, transforming how people interact, work, and play. Users require a head-mounted display (HMD), software platforms, services, and network connectivity to experience the Metaverse, making exceptional network traffic management (NTM) essential for Internet Service Providers (ISPs) to deliver optimal Quality of Service (QoS) and Quality of Experience (QoE) \cite{wang2023survey}. Low latency and adequate bandwidth are crucial for Metavrese to avoid cybersickness \cite{10124955}. Therefore, efficient resource allocation and network tuning are vital for ISPs to manage the increasing demand, which requires accurate identification of Metaverse network traffic through network traffic classification (NTC) \cite{Futurene65}. Identifying Metaverse network traffic serves multiple purposes, including traffic prioritization, resource allocation, security, and monetization \cite{wang2023survey}. Also, the NTC is required to optimize the QoS, UE route selection policy (URSP), and for security, highlighting the need for further research in this area \cite{IETF3924}, especially for the unexplored Metaverse network traffic.

A decision trees-based AR and cloud gaming traffic classification is proposed in \cite{shirmarz2024pixels}. The work also provides pre-processed uplink and downlink AR traffic data. While, on average, the work achieves 96\% accuracy in classifying uplink traffic, the accuracy in classifying downlink traffic is limited on 89\%. Downlink traffic is bandwidth-demanding, and the classification accuracy achieved by this work is not sufficient to help efficient traffic management. Other work \cite{manjunath2022segmented} helps classify VR traffic using application-level information, which is imperative in avoiding expensive deep packet inspection (DPI) software that violates privacy-related policies. The solution provides stellar 99\% accuracy in classifying VR traffic among other applications. However, it is difficult for the proposed method to be further generalized to classify the traffic of other Metaverse services, such as AR, MR, and other VR-related services. Recently, a deep learning-based solution is proposed for Metaverse traffic classification \cite{10579124}. The solution achieves 87\% accuracy while classifying the traffic of three Metaverse classes: network infrastructure, real-time conversation, and non-conversational applications. Based on the literature survey, we identify the following research gaps: i) non-availability of comprehensive real-world Metaverse network traffic data, and ii) accurate Metaverse network traffic classifer. We address both gaps in this paper. The work \cite{10579124} is considered a state-of-the-art work (SoA) since it is close to our experiment; however, the data used in the work are not pure Metaverse network traffic.



We propose a segmented learning-based Metaverse network traffic classifer, called Discern-XR. Our solution treats the Metaverse network traffic in segments of traffic packets with four raw features: time, packet length, packet direction, and packet inter-arrival time. We propose a Frame Vector Representation (FVR) algorithm that uses the Frame Identification Algorithm (FIA) to extract frame-related statistics and the statistics of the segments from the four raw features. The working principle behind the proposed algorithm is to identify the different statistical behaviours in the Metaverse services that provide vital and distinctive information for a superior classification. We propose an Augmentation, Aggregation, and Retention Online Training (A2R-(OT)) algorithm that converges to find the optimal segment size to align with the working principle and finds classification model through online training. Our solution outperforms all Metaverse traffic classifiers by 7\% while reducing the training time for a quick decision. 

Our work contributes the following to the Metaverse network traffic classification: (i) a real-world, holistic, and comprehensive Metaverse network traffic data that consists of VR Games, VR Video, VR Chat/VoIP, AR, and MR traffic flows, available in \cite{data_1}, (ii) a state-of-the-art frame vector representation (FVR) and video frame identification algorithm (FIA) for Metaverse services, (iii) a state-of-the-art Augmentation, Aggregation, and Retention-Online training (A2R-(OT)) algorithm to find an accurate classifier that uses application-level features and (iv) implementation of the solution is made open source at \cite{yogasuha5}. The Discren-XR can be employed at any stage of network infrastructure since it uses application-level features. Therefore, our solution benefits ISPs and network router manufacturers by enhancing QoS for Metaverse-related services. 

\section{System model}
\label{sec:sm}

Let $\bm{p}_{j,i}$ represent the packet of a Metaverse traffic service $\bm{s}_i$, where $i$ is an index for different services and $j$ is an index of packets. Each packet $\bm{p}$ is a vector with four raw features: time, packet length, packet direction, and packet inter-arrival time. Let $\bm{X} = \{\bm{X}_1, \bm{X}_2,...,\bm{X}_{N}\}$ represent the set of $N$ network traffic segments. The element $\bm{X}_k$, for $k = \{1,2,...,N\}$, is a matrix of dimension ($S\times4$), where $S$ is the size of the segment, and 4 is the number of features. The raw network traffic segment $\bm{X}_k$ is transformed into a set of statistical feature vectors $\bm{v}(\bm{X}_k)$ through a feature transformation function $\phi$, which transforms the raw matrix into a statistical feature vector, which is given as
\begin{equation}
    \label{eq:1}
    \bm{v}(\bm{X}_k) = \phi(\bm{X}_k).
\end{equation}
After integrating the feature vectors for all traffic segments, the resultant feature matrix is given as $\bm{V} = \{\bm{v}(\bm{X}_1), \bm{v}(\bm{X}_2),\dots,\bm{v}(\bm{X}_N)\}$. Each statistical vector $\bm{v}(\bm{X}_k)$ is associated with a service label $\bm{y} \in \{1,2,3,...,C\}$, where $C$ represents the number of services. 
A mapping function $F$ classifies each segment into its correct service category, which is given as
\begin{equation}
    \label{eq:2}
    F(\bm{v}(\bm{X}_k)) = y_k.
\end{equation}
We need to maximize the accuracy while finding a mapping function $F$ by minimizing the error. Let $\mathcal{L}(y_k, \hat{y}_k)$ be the loss function that quantifies the error between the true label $y_k$ and the predicted label $\hat{y}_k$. The training objective is 
\begin{equation}
    \label{eq:3}
    \min_F \sum_{k=1}^{N}\mathcal{L}(y_k, \hat{y}_k),
\end{equation}
where $F$ represents the mapping function and $N$ represents the number of segments. Data, segment size, and statistical features during the training play a crucial role in minimizing the loss function.

The proposed solution operates on two principles: the similarity of statistical characteristics for the traffic segments from the same service and the uniqueness of these statistical patterns across different services. 

\textbf{Principle 1}: Vector $\bm{s}$ represents a set of different services, where $i$ is the index of services. Vector $\bm{v}_i^k$ denotes the statistical vector of the $k$-th traffic segment of $i$-th service corresponding to $s_i$. For all statistical vectors of segments, $\bm{v}_i^k$, within a service $s_i$, the statistical features should follow a similar distribution, denoted as $f(s_i)$. The property is given as
$$f(\bm{v}_i^k) \approx f(s_i), \forall k,$$
indicating that the segments from the same service exhibit statistical similarity.

\textbf{Principle 2}: 
Let $s_i$ and $s_j$ be different services, where $i \neq j$. According to principle 1, a similar statistical distribution is expected among the segments for each service. However, the statistical distributions of those segments for two services are different, which is given as
$$ f(s_i)\neq f(s_j).$$



Note that Principle 1 holds well with the optimal segment size to capture statistical similarities. If the size of segments is too small, the statistical features might not capture the underlying distribution of the service; if it is too large, the data may be overly generalized, losing service-specific patterns. By nature, different Metaverse services exhibit unique statistical behaviour for Principle 2 to work. 

\section{Discern-XR: Metaverse Network Traffic Classifcaiton}
\label{sec:SegFrame}



\begin{figure}
 	\centering
 	\includegraphics[trim={0 2 2 0},width=\linewidth]{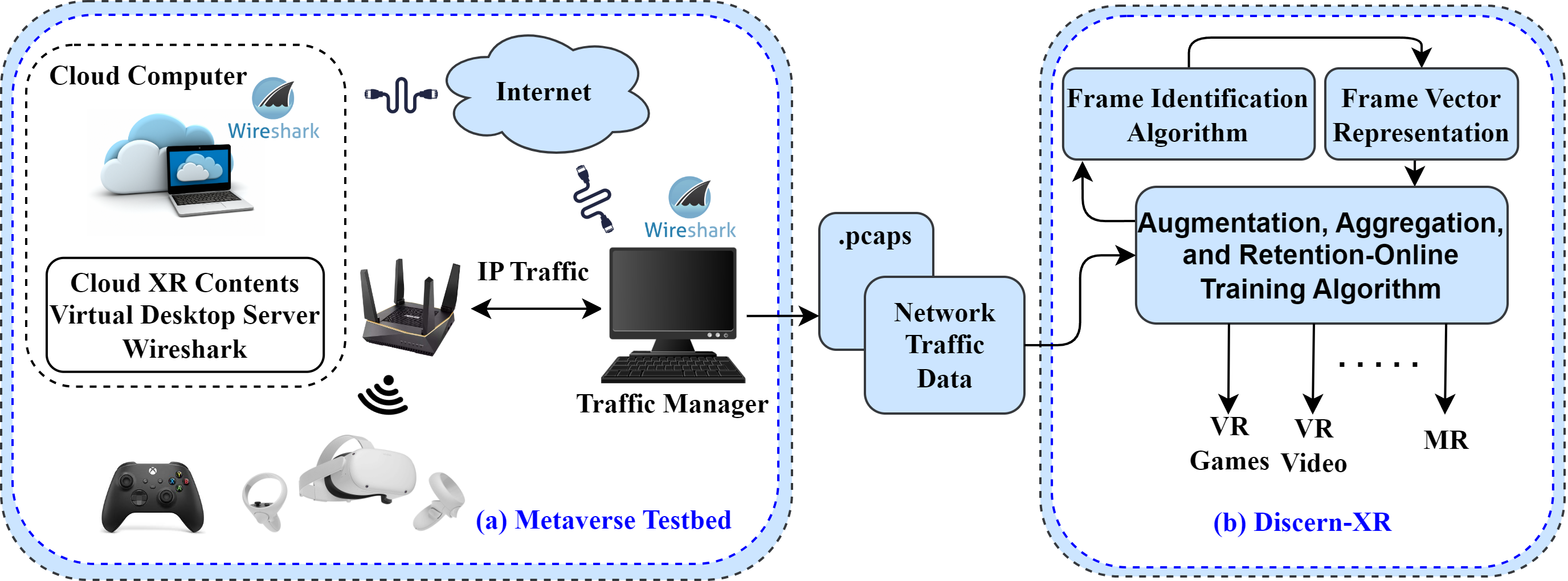} 
 	\caption{Overview of the proposed solution. (a) Metaverse testbed to capture Metaverse network traffic, and (b) block diagram of the Discern-XR solution.} 

 	\label{fig:sm}
\end{figure}

We select diverse and popular Metaverse services cloud-rendered to an Oculus Quest 2 HMD \cite{MetaQues31} using Virtual Desktop Streamer (VDS) \cite{HomeVirt30}. The rendered traffic is tapped on a cloud computer using a traffic sniffer, i.e., Wireshark \cite{wireshark}. Wireshark extracts the captured traffic in packet captures (.pcap) files from which network traffic data is extracted into comma-separated values (CSV). The extracted CSV for a given service consists of four application-level features. The structural components of the Metaverse testbed is shown in Figure \ref{fig:sm} (a). The devised Metaverse traffic classifier, Discren-XR receives the Metaverse network traffic at the A2R-(OT) that invokes the FVR and FIA with the required segment size to form statistical frame vectors that are used in finding the classification model in training. Once the A2R-(OT) finds the accurate classifier, the training is stopped, and the learned model is deployed for further classification as shwon in Figure \ref{fig:sm} (b).

\subsection{Frame Identification Algorithm}
\label{sec:fia}

\begin{figure}
\title{}
\centering
\subfloat[]{%
  \includegraphics[width=0.8\linewidth]{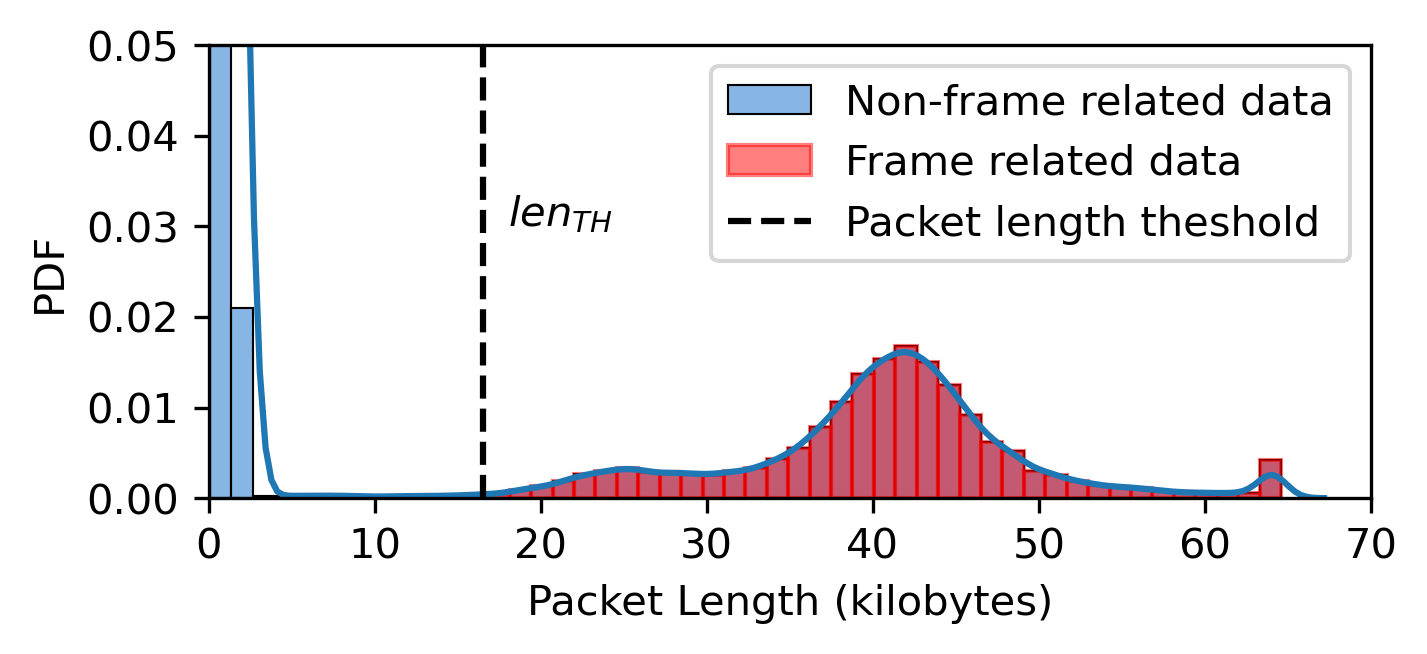}%
  \label{fig:iat+pktlen(a)}
}
\vspace{-0.01cm}
\subfloat[]{
\includegraphics[width=0.8\linewidth]{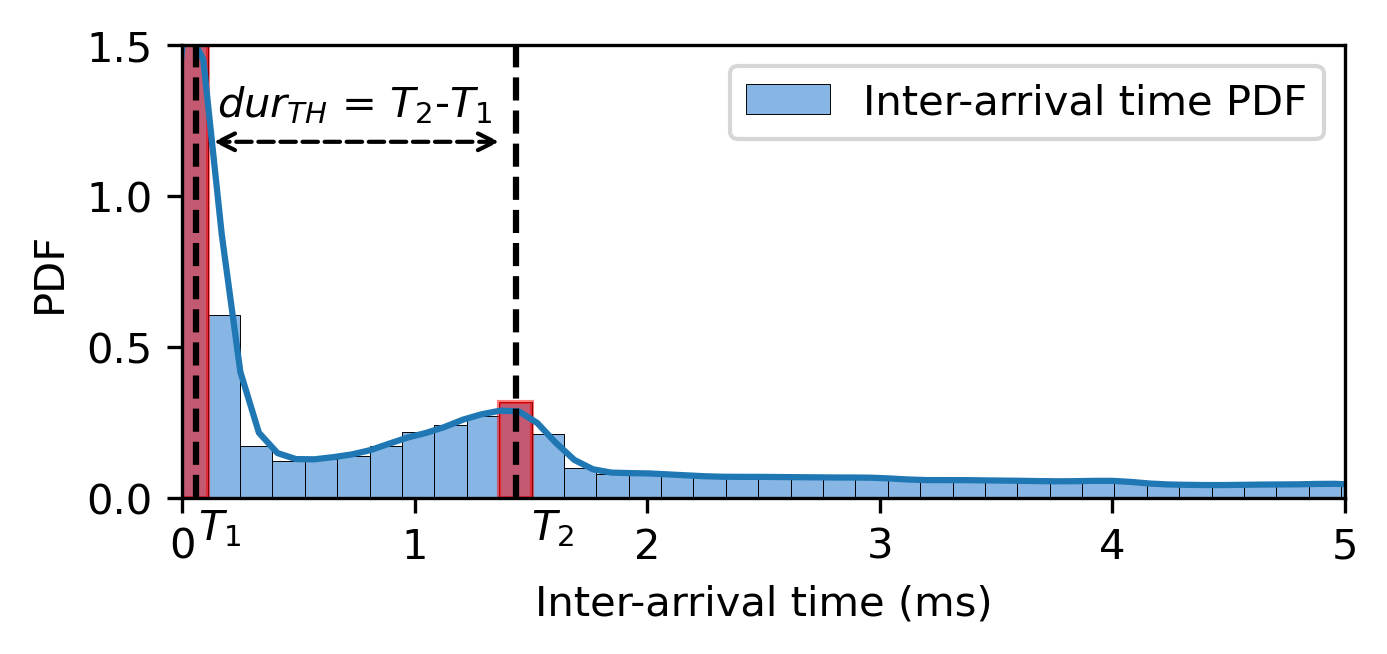}%
\label{fig:iat+pktlen(b)}
}
\caption{(a) PDF of packet lengths, and (b) PDF of inter-arrival time for a sample Metaverse traffic segment. }.
\label{fig:iat+pktlen}
\end{figure}


 



 


    
        
    
        
        
 
 

Metaverse network traffic is significant in the downlink direction while rendering video and audio frames. Insignificant uplink traffic consists of control flow generated from sensors/joystick at the HMD end \cite{9685808}. The patterns of video frames provide unique information about the type of Metaverse services. Therefore, identifying frames regardless of rendering platforms can be crucial in Metaverse traffic classification. The FIA algorithm relies on the traffic behaviour, including packet length and inter-arrival time, to accurately identify video frames. This is because multiple consecutive packets are often required to transmit a relatively large, uniform frame-related video traffic compared to non-frame traffic. The flow of frame-related video traffic is similar and relatively large compared to non-frame-related traffic flow. In addition, packets related to the same frame are sent consecutively and in quick succession. The disparity in packet length allows the algorithm to define a minimum packet length threshold for identifying frames as depicted in Figure \ref{fig:iat+pktlen(a)}. The reliability in frame packet inter-arrival times allows the algorithm to define the maximum frame duration as the difference in mode inter-arrival times. As illustrated in Figure \ref{fig:iat+pktlen(b)}, the first mode ($T_1$) represents video and acknowledgement packets to the video traffic flow, whereas the second mode ($T_2$) represents audio and control traffic flow. In Figure \ref{fig:iat+pktlen} it is shown that $len_{TH}$ is determined as 25\% of the maximum length of the observed packet length. $dur_{th}$ is the frame duration threshold determined between the first two peaks. The first peak represents the start of the video frame packet with less inter-arrival time, and the second peak represents the end of the video frame. The FIA algorithm uses this to guarantee that packets with significant inter-arrival times are not considered frames-related traffic flow and to ensure that multiple transmitted frames are not identified as single frames. 

\subsection{Frame Vector Representation}
\label{sec:fvr}


 
 
  
 
    
    
    

    

    
    

\begin{figure}
	\centering
	\includegraphics[trim=0 0 0 0 ,width=0.9\linewidth]{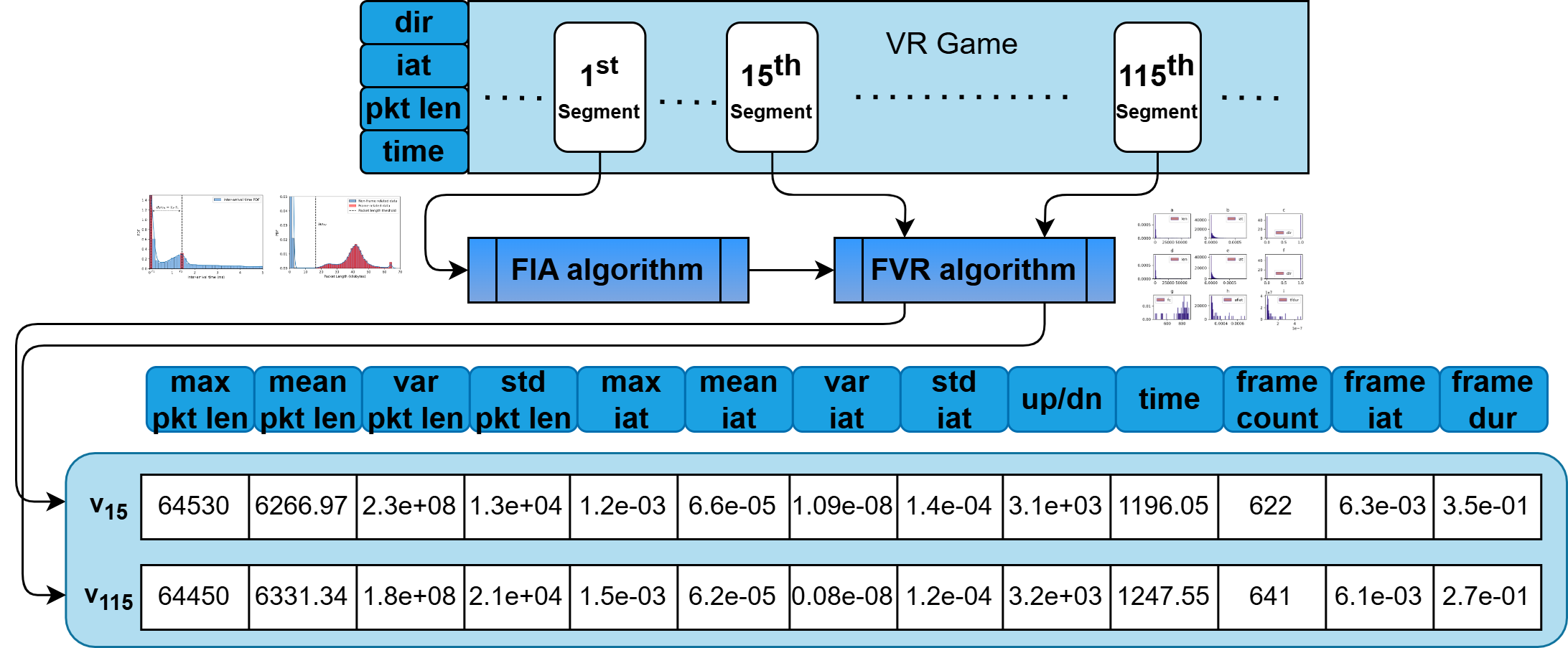}
	\caption{Vectorization of traffic segment using FVR algorithm. Representing statistical frame vector $\bm{v}_i$ for the $i$\textsuperscript{th} segment. Diagram shows the representation of FVR for 15\textsuperscript{th} and 115\textsuperscript{th} segment from VR Game.}
	\label{fig:fvr}
\end{figure}

\begin{figure}
\title{}
\centering
\subfloat[]{%
  \includegraphics[width=0.8\linewidth]{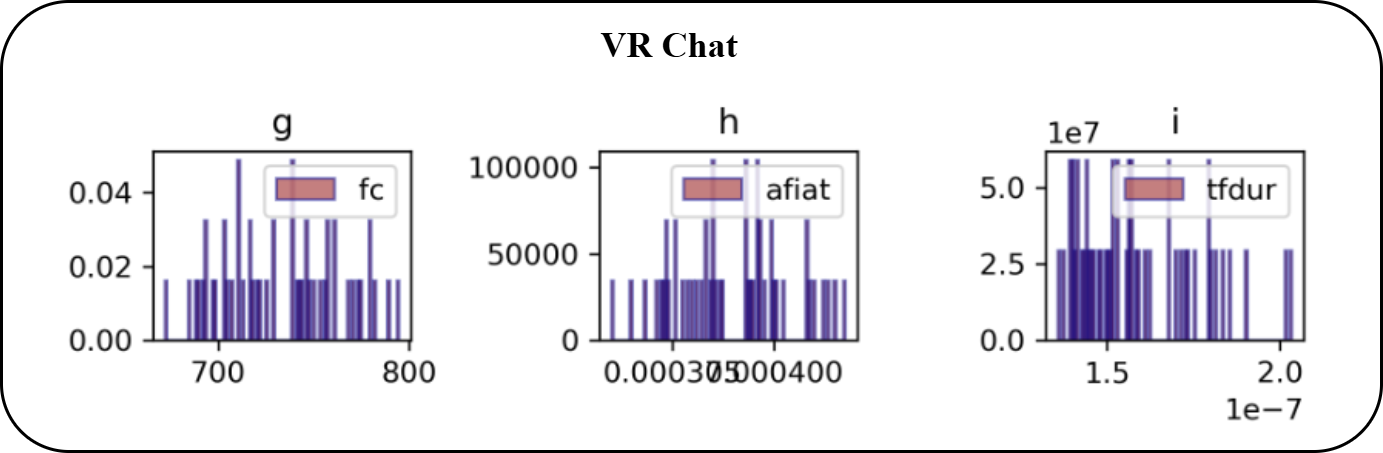}%
  \label{fig:fvr_chat}
}
\vspace{-0.01cm}
\subfloat[]{
\includegraphics[width=0.8\linewidth]{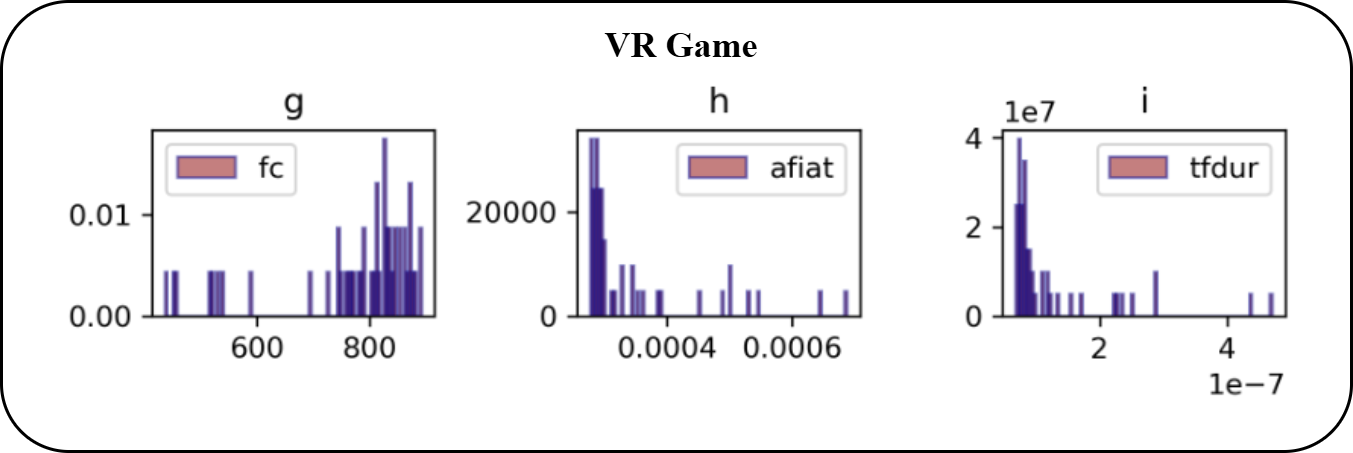}%
\label{fig:fvr_game}
}
\caption{(a) Histogram of frame count (g), average frame inter-arrival time (h), total frame duration (i) for 115th segment of VR Chat service, and (b) Histogram of frame count (g), average frame inter-arrival time (h), total frame duration (i) for 115th segment of VR Game service. Chat service shows many small frames (g) with scattered inter-arrival time (h) and duration (i). Game service shows larger frames (g) with many packets with small inter-arrival time (h) and smaller duration (i). Therefore, frame-related information provides unique information for superior classification.}
\label{fig:VRCG}
\end{figure}


The FVR algorithm represents a given traffic segment into a statistical frame vector $\bm{v}_i$, which contains 13 statistical features derived from the four raw features, as shown in Figure \ref{fig:fvr}. The first ten features are related to the statistical information of the raw traffic data, which provides holistic information on traffic behaviour. The final three features are derived from the frame-related traffic data: frame count, average frame inter-arrival time, and total frame duration, which provide unique information about Metaverse traffic services, as shown in Figure \ref{fig:VRCG}. 

\subsection{Augmentation, Aggregation, and Retention-Online Training Algorithm}
\label{sec:A2R_OT}

\begin{algorithm}
\caption{A2R-(OT) and segment size selection}\label{alg:A2R}
\KwData{Metaverse network traffic data}
\KwResult{\textit{final model}, $S_T$ ,$S$}
 \textbf{initialization}\;
 
$T_{i}$ = 50; $Z_{error}$ = $0$; $E_{stop}$ = $0$; models = $[$ $]$; data = $[$ $]\;$
  
 \While{$S_{i}$ != $S_{max}$}{
    \While{$N_f$ != $N_{opt}$}{
         data.append($S_{i}$)
        
         trainData = FVR$\left(S_{i},N_{f},Network \;data\right)$
         
         valData = split(trainData, $V_{R}$) \Comment{split() will split the trainData at $V_{R}$, which is a validation ratio;}
         
         model = RandomForest((50x$T_{i}$),trainData)
    
         models.append(model)
         
         ${E}_{curr}$ = model.test(valData)
    
         $S_{i}++$
         
         \While{($E_{stop}$ $<$ $ES_{TH}$) and ($Z_{error}$ $<$ $ZE_{TH}$ )}{
                currentError = ${E}_{curr}$   
    
                data.append($S_{i}$)
    
                trainData = FVR$\left(S_{i},N_{f},Network \;data\right)$
         
                valData = split(trainData, $V_{R}$)
         
                model = RandomForest((50x$T_{i}$),trainData)
    
                models.append(model)
         
                error = model.test(valData)
    
                $S_{i}++$
                
                \If{${E}_{curr}$ $==$ $0$}{
                    $Z_{error}++$
                }

                $\Delta E=$ currentError - error
                
                \If{abosluteValue($\Delta E$) $\leq$ $E_{TH}$}{
                    $E_{stop}++$
                }
         }

         $N_f \mathrel{+}= 500$
     }

    trainData = FVR$\left(data,N_{f},Network \;data\right)$

    valData = split(trainData, $V_{r}$)

    model = RandomForest((50x$T_{i}$),trainData)

    models.append(model)

    $S$ = $N_f$; $S_T$ = $S_{i}$
       
    \textit{final model} = combine(models) \Comment{concatenate all trained models;}

    
  }
\Return {final model, $S_T$ ,$S$}
\label{alg:a2r}
\end{algorithm}

The proposed A2R-(OT) algorithm, presented in Algorithm \ref{alg:a2r}, adopts the random forest algorithm, which continuously refines the Metaverse classifier by iterating through various segment sizes to find the optimal segment size ($S$), number of training segments ($S_T$), and final classification model (\textit{final model}). The outer loop determines the number of training segments ($S_T$), while the inner loop refines the segment size ($S$). The algorithm start by forming segment. The FVR forms the vectors of the respective segments. Split function helps splitting the segment vectors into train and validation data at ration $V_R$. Random forest is trained with train data until the validation meet the stopping criteria: 1) zero error conditions and 2) early stopping conditions. As given in Algorithm \ref{alg:A2R}, zero error flag (${z}_{error}$) is incremented when the current error (${E}_{curr}$) is $0$. Similarly if the change in the error ($\Delta {E}$) is less than error threshold (${E}_{TH}$), early stop flag (${E}_{stop}$) is incremented. The optimization process is stopped when one of the conditions is met: 1) ${z}_{error} \geq {ZE}_{TH},$ and 2) ${E}_{stop} \geq {ES}_{TH}.$ 

The objective function of the A2R-(OT) algorithm is given in Eq.\eqref{eq:3}. 
  Hyperparameters of random forest affect the optimization process. The classification model's error can be minimized by reducing the variance by increasing the number of trees ($T_{RF}$). In other words, mathematically given as $\text{Accuracy} \propto \sqrt{T_{RF}}$ \cite{biau2016random}. Warm-start is enforced to increase the trees' depth. However, we will use smaller segments during the training to avoid overfitting. The time complexity of the A2R-(OT) algorithm is approximately $O\left(S_{\text{max}} * \left(\frac{N_{\text{opt}}}{\Delta N_f}\right) * T * N_f * \log(N_f)\right)$, where $S_{\text{max}}$ represents the total number of segment sizes, $N_{\text{opt}}$ is the optimal segment size, $\Delta N_f$ is the increment in segment size, $T$ is the number of trees in the random forest, and $N_f$ is the segment size. Random forest training is the most computationally expensive part of this process, especially as the segment size $N_f$ increases with higher dynamic behaviour.

The A2R-(OT) algorithm operates on three core principles: Augmentation, where new network traffic segments are continuously added to improve generalization; Aggregation, where multiple models trained on different segments are combined for a more robust final model; and Retention, which ensures the model retains and builds on previous knowledge in dynamic environments like Metaverse traffic, ensuring sustained accuracy and efficiency.

\section{Experimentation Setup and Results}

\subsection{Datasets, Implementation, and plan of experiments}
The dataset from our testbed is available in \cite{data_1}. We use the datasets Dataset I, \cite{data_1}, Dataset II, \cite{9685808}, and Dataset III, \cite{questset}. 
Table \ref{table:exppss} shows the experimental plan. Datasets I, II, and III are used in all experiments. Dataset III provides two different data labels, as experiments 3 and 4 show in the Table \ref{table:exppss}. In experiment 5, we use Dataset III in the training and Dataset II for testing to explore the generalization of our solution. We separately train the model for each service. Once the algorithm finishes converging, the remaining data is used for testing.  Each of the above mentioned experiments is executed five times to check the robustness of the A2R-(OT) algorithm's convergence. The FVR algorithm finds the weight $\phi$ in Eq. \eqref{eq:1}. 
Initial segment size ($N_f$) in Algorithm \ref{alg:a2r} is set to 500 and increments with 500 packets per iteration; the number is selected because of the dynamic nature of Metaverse services. smaller numbers might not help capture statistical similarities explained in Section \ref{sec:sm}. 
For our experiment, $S_{max}$ is capped at 200 to avoid data exhaustion, and training continues until the minimum error threshold of 2\% or zero error is met. During the empirical test, we noticed the minimum error was 2\%, therefore the threshold is determined at this value. Some segments provide holistic information that helps to converge to zero error. On average, we use 40\% of the data in training, 20\% of the data for validation, and 40\% of the data for testing. 





The solution is implemented in Python using data science libraries such as Sklearn, NumPy and Pandas. The implementation of the solution is available at \cite{yogasuha5}. The experiment is conducted on a Windows platform with an NVIDIA RTX2800S graphical processing unit. The Windows platform is installed with the Anaconda environment to necessitate ML-related libraries to run along with TensorFlow.





\begin{table}[h]
\caption{Datasets and Experimental plan.}
\scriptsize
\centering 
\begin{tabular}{
>{\centering\arraybackslash}m{1.5cm} 
>{\centering\arraybackslash}m{1.2cm}
>{\centering\arraybackslash}m{1.2cm} 
>{\centering\arraybackslash}m{3cm}} 
\hline
Experiment Number& Train Dataset & Test Datset & Multi-Class Label \\
\hline 
\multirow{1}{4em}1 & \rotatebox[origin=c]{0}{{Dataset I }} & \rotatebox[origin=c]{0}{{Dataset I }}  & \Longstack{ VR Video \\  VR Game \\VR Chat\\ Augmented Reality\\ Mixed Reality  }\\

\hline
\hline
\multirow{1}{4em}2 & \rotatebox[origin=c]{0}{Dataset II} &  \rotatebox[origin=c]{0}{Dataset II}& \Longstack{ VR Fast Traffic \\  VR Slow Traffic}\\
\hline
\hline
\multirow{1}{4em}3 & \rotatebox[origin=c]{0}{Dataset III} & \rotatebox[origin=c]{0}{Dataset III}  & \Longstack{ VR Fast Traffic Game 1 \\ VR Fast Traffic Game 2\\ VR Slow Traffic Game 3 \\ VR Slow Traffic Game 4}\\
\hline
\hline
\multirow{1}{4em}4 &\rotatebox[origin=c]{0}{Dataset III} & \rotatebox[origin=c]{0}{Dataset III} & \Longstack{ VR Slow Traffic \\ VR Fast Traffic} \\
\hline
\hline
\multirow{1}{4em}5 &\rotatebox[origin=c]{0}{Dataset III} & \rotatebox[origin=c]{0}{Dataset II} & \Longstack{ VR Slow Traffic \\ VR Fast Traffic} \\

\hline
\end{tabular}
\label{table:exppss}
\end{table}

\subsection{Performance Metrics}
\label{pm}

We use Accuracy, Recall, Precision, F1 score, and False Negative Rate (FNR) to evaluate the classification model. These metrics are defined based on True Positives (TP), True Negatives (TN), False Positives (FP), and False Negatives (FN). Accuracy measures the proportion of correct predictions, calculated as 

\begin{equation}
    \text{Accuracy} = \frac{TP + TN}{TP + TN + FP + FN}.
\end{equation}
Recall, or Sensitivity, is the ratio of correctly predicted positive instances, given by 
\begin{equation}
    \text{Recall} = \frac{TP}{TP + FN}.
\end{equation}
Precision evaluates the accuracy of positive predictions, expressed as 
\begin{equation}
    \text{Precision} = \frac{TP}{TP + FP}.
\end{equation}
The F1 score is the harmonic mean of Precision and Recall, providing a balance between the two, and is defined as 
\begin{equation}
    \text{F\_1} = 2 \times \frac{\text{Precision} \times \text{Recall}}{\text{Precision} + \text{Recall}}. 
\end{equation}Finally, the False Negative Rate (FNR), which is critical in Network Traffic Classification (NTC) problems, represents the proportion of missed positive instances and is given by 
\begin{equation}
    \text{FNR} = \frac{FN}{TP + FN}.
\end{equation}
A model with a higher F1 score is generally considered robust, but reducing the FNR is crucial, as it indicates situations where the model fails to detect actual traffic. Ideally, a classification model should aim for a high F1 score and a low FNR for better reliability and accuracy.




\subsection{Performance of Frame Identification Algorithm}
Table \ref{table:fiap} provides the frame rate from the FIA for 60 Hz. We also verified the results with the VDS readings during the data capture and found the findings accurate. Ten thousand bytes is the packet threshold, and the inter-arrival time threshold for each service is given in the second column of Table \ref{table:fiap}. Please refer to Section \ref{sec:fia} for the information on the packet and the inter-arrival time threshold required for FIA. The FIA provides accurate results for all services except for VR chat. The asynchronicity from VR chat poses a challenge for an accurate frame identification in our solution.

\begin{table}[h]
\centering
\scriptsize
\caption{Frame rate per second from FIA algorithm.}
\begin{tabular}{|l|c|c|c|}
\hline
\textbf{Application} & \textbf{Frame Rate (Hz)} & \textbf{IAT} (s) & \textbf{\% error} (\%)\\ \hline


\multicolumn{4}{|c|}{\textbf{60 Hz}} \\ \hline
MR      & 59.85  & 0.016 & 0.25\\ 
\textbf{AR}       & \textbf{60.00}   & 0.016 & \textbf{0}\\ 
VR Video     & 59.74  & 0.017 &0.43\\
VR Game     & 60.01 & 0.016 & 0.016 \\
VR Chat        & 68.28  & 0.014 & 13.8 \\  \hline
\end{tabular}
\label{table:fiap}
\end{table}
\subsection{Performance of the A2R-(OT) algorithm}

\begin{figure}
	\centering
	\includegraphics[width=\linewidth]{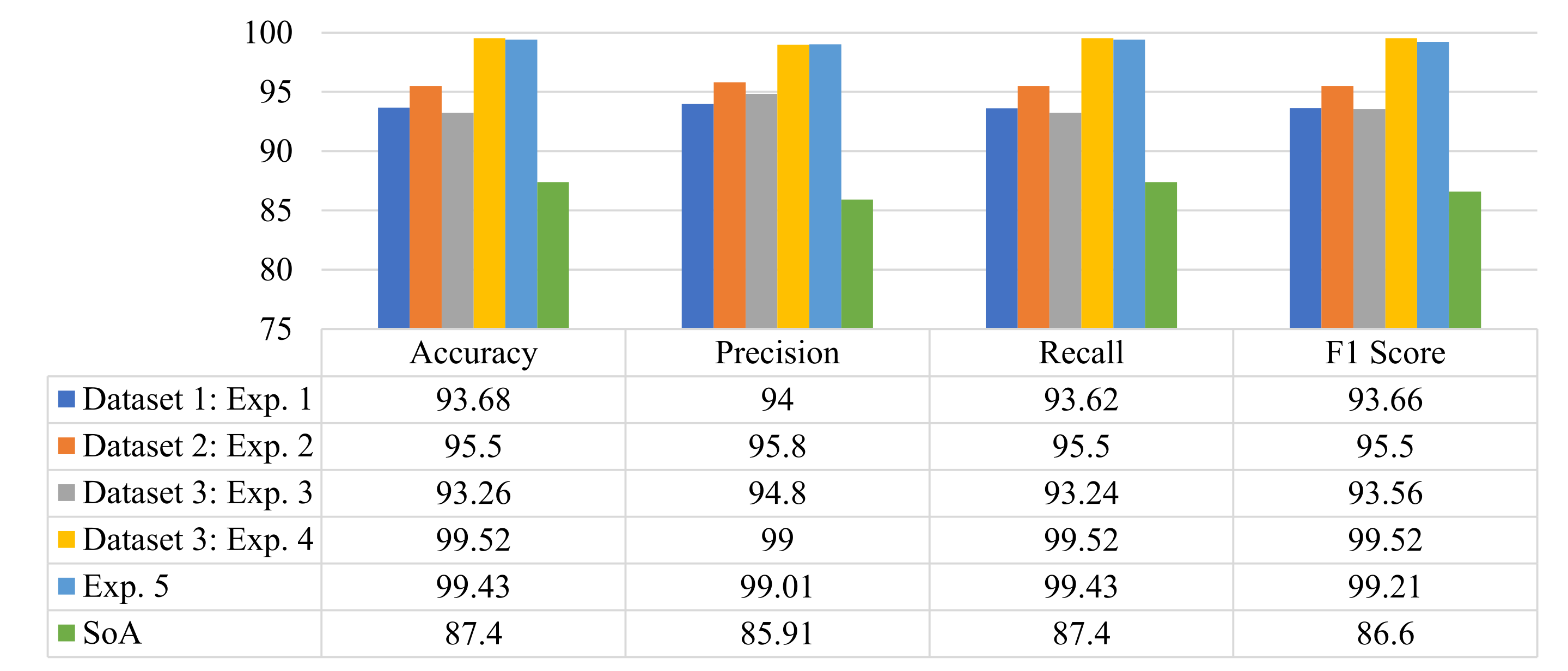}
	\caption{Results for the multi-class classifier for different experiments given Table \ref{table:exppss}.}
	\label{fig:d2_p1}
\end{figure}

Figure \ref{fig:d2_p1} shows the performance of our solution for various experiments presented in Table \ref{table:exppss}. The solution consistently produces accuracy higher than 93\% for in-house and public datasets. We provide accuracy and FNR per service from each experiment in Table \ref{table:mcct} along with a number of test segments utilized in the experiments. FNR is smaller in all cases. However, VR games, AR, and MR services show less accuracy because of higher dynamicity and minor inconsistency with the FIA algorithm when considering the traffic in segments. Overall, the accuracy and FNR for all services are satisfactory and 7\% better than the SoA \cite{10579124} . 

\subsection{Discussions}
\begin{figure}
	\centering
	\includegraphics[width=0.8\linewidth]{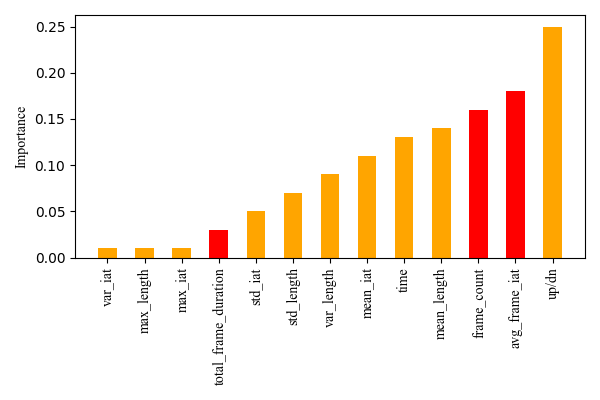}
	\caption{Feature importance of random forest for Experiment 1 given Table \ref{table:exppss}.}
	\label{fig:fi}
\end{figure}

In each experiment, the A2R-(OT) algorithm plays a crucial role in determining the appropriate segment size and the number of segments used for training, as shown in Table \ref {table:mcct1}. The highly dynamic nature of Metaverse traffic is evident in the large segment size. However, the A2R-(OT) converges faster with fewer segments, which aligns with our expectations to reduce training time. The Metaverse traffic is highly random at the start of the session, and unless the user or network health introduces uncertainty, it remains predictable. Figure \ref{fig:fi} shows the frame-related information, shown in red, provides better information for the classification.

\begin{table}
\caption{Accuracy and FNR for different experiments.}
\scriptsize
\centering

\begin{tabular}{
>{\centering\arraybackslash}m{0.4cm} 
>{\centering\arraybackslash}m{2.7cm} 
>{\centering\arraybackslash}m{1.4cm}
>{\centering\arraybackslash}m{1cm} 
>{\centering\arraybackslash}m{1cm}}
\hline
&CoS label & Number of test segments & Accuracy (\%) & FNR \\
\hline 
\multirow{5}{4em}{\Longstack{\rotatebox[origin=c]{0}{Exp. 1} }}&VR Video & 68 & 98.53 & 0.0147\\
&VR Game & 89 & 84.27 & 0.015\\
&\textbf{VR Chat/VoIP} & 53 & \textbf{100} & \textbf{0}\\
&AR & 79 & 92.41 & 0.07\\
&MR & 76 & 92.11 & 0.07\\
\hline
\hline
\multirow{2}{4em}{\rotatebox[origin=c]{0}{Exp. 2}} 
&\textbf{VR Fast Traffic}&	132&	\textbf{95.52}&	{0.075}\\
&VR Slow Traffic&	114&	95&	\textbf{0.008}\\
\hline
\hline

\multirow{4}{4em}{\Longstack{\rotatebox[origin=c]{0}{Exp. 3} }}&VR Fast Traffic Game 1 & 257 & 94.3 & 0.017\\
&VR Fast Traffic Game 2 & 123 & 95.69 & 0.016\\
&\textbf{VR Slow Traffic Game 3}  & 350 & \textbf{99.65} & \textbf{0.008}\\
&VR Slow Traffic Game 4 & 130 & 96.62 & 0.06\\
\hline
\hline

\multirow{2}{4em}{\Longstack{\rotatebox[origin=c]{0}{Exp. 4} }}&\textbf{VR Fast Traffic}& 847 & \textbf{99.5} & \textbf{0.001}\\
&VR Slow Traffic & 980 & 99.5& 0.008\\
\hline
\hline

\multirow{2}{4em}{\Longstack{\rotatebox[origin=c]{0}{Exp. 5} }}&VR Fast Traffic& 172 & 98.6 & \textbf{0.005}\\
&\textbf{VR Slow Traffic} & 154 & \textbf{99.1}& 0.006\\
\hline
\hline

\end{tabular}
\label{table:mcct}
\end{table}

\begin{table}
\centering
\scriptsize
\caption{Number of segments and size used for training.}
\begin{tabular}{
>{\centering\arraybackslash}m{1cm} 
>{\centering\arraybackslash}m{2cm} 
>{\centering\arraybackslash}m{2cm}
>{\centering\arraybackslash}m{2cm}}
\hline
\textbf{Experiment} & \textbf{Segment Size} & \textbf{No. Segments used for Training}  & \textbf{Training Time (sec)}\\ \hline
\multirow{1}{*}{\rotatebox[origin=c]{0}{Exp. 1}} & 6000  & 12 & 92\\ \hline
\multirow{1}{*}{\rotatebox[origin=c]{0}{Exp. 2}} & 6000  & 10 & 76\\ \hline
\multirow{1}{*}{\rotatebox[origin=c]{0}{Exp. 3}} & 16000 & 9 & 165\\ \hline
\multirow{1}{*}{\rotatebox[origin=c]{0}{Exp. 4}} & 12000 & 11 & 147\\ \hline
\multirow{1}{*}{\rotatebox[origin=c]{0}{Exp. 5}} & 12000 & 9 & 154\\ \hline
\end{tabular}
\label{table:mcct1}
\end{table}

\section{Conclusion and Future work}

We have proposed the Discenr-XR classification framework for Metaverse network, i.e., Discern-XR,  to identify VR, AR, and MR service traffics. Using our SoA algorithms, FIA, FVR, and A2R-(OT), the framework demonstrates superior accuracy and performance, improving the detection of Metaverse-related traffic by 7\% compared to existing methods while reducing the modelling time. Discern-XR can play a crucial role in the rapidly evolving Metaverse environment, enhancing the ability of ISPs to manage traffic efficiently and improving QoS and QoE for users.
Future efforts will further extend Discern-XR’s scalability to accommodate the increasing variety of Metaverse services. The framework will be adapted for next-generation networks, such as 5G and beyond, where low latency and high bandwidth are critical.

\bibliographystyle{IEEEtran}
\bibliography{refs}

\begin{thebibliography}{10}
\providecommand{\url}[1]{#1}
\csname url@samestyle\endcsname
\providecommand{\newblock}{\relax}
\providecommand{\bibinfo}[2]{#2}
\providecommand{\BIBentrySTDinterwordspacing}{\spaceskip=0pt\relax}
\providecommand{\BIBentryALTinterwordstretchfactor}{4}
\providecommand{\BIBentryALTinterwordspacing}{\spaceskip=\fontdimen2\font plus
\BIBentryALTinterwordstretchfactor\fontdimen3\font minus \fontdimen4\font\relax}
\providecommand{\BIBforeignlanguage}[2]{{%
\expandafter\ifx\csname l@#1\endcsname\relax
\typeout{** WARNING: IEEEtran.bst: No hyphenation pattern has been}%
\typeout{** loaded for the language `#1'. Using the pattern for}%
\typeout{** the default language instead.}%
\else
\language=\csname l@#1\endcsname
\fi
#2}}
\providecommand{\BIBdecl}{\relax}
\BIBdecl

\bibitem{wang2023survey}
H.~Wang, H.~Ning, Y.~Lin, W.~Wang, S.~Dhelim, F.~Farha, J.~Ding, and M.~Daneshmand, ``A survey on the metaverse: The state-of-the-art, technologies, applications, and challenges,'' \emph{IEEE Internet of Things Journal}, vol.~10, no.~16, pp. 14\,671--14\,688, 2023.

\bibitem{10124955}
M.~Li, J.~Gao, C.~Zhou, X.~Shen, and W.~Zhuang, ``User dynamics-aware edge caching and computing for mobile virtual reality,'' \emph{IEEE Journal of Selected Topics in Signal Processing}, vol.~17, no.~5, pp. 1131--1146, 2023.

\bibitem{Futurene65}
``Future network trends driving universal metaverse mobility - ericsson,'' \url{https://www.ericsson.com/en/reports-and-papers/ericsson-technology-review/articles/technology-trends-2022}, (Accessed on 08/15/2024).

\bibitem{IETF3924}
F.~Baker, B.~Foster, and C.~Sharp, ``The {Cisco} architecture for lawful intercept in {IP} networks,'' Available at \url{https://tools.ietf.org/html/rfc3924}.

\bibitem{shirmarz2024pixels}
A.~Shirmarz, F.~L. Verdi, S.~K. Singh, and C.~E. Rothenberg, ``From pixels to packets: Traffic classification of augmented reality and cloud gaming,'' in \emph{2024 IEEE 10th International Conference on Network Softwarization (NetSoft)}.\hskip 1em plus 0.5em minus 0.4em\relax IEEE, 2024, pp. 195--203.

\bibitem{manjunath2022segmented}
Y.~S.~K. Manjunath, S.~Zhao, H.~Abou-zeid, A.~B. Sediq, R.~Atawia, and X.-P. Zhang, ``Segmented learning for class-of-service network traffic classification,'' in \emph{GLOBECOM 2022-2022 IEEE Global Communications Conference}.\hskip 1em plus 0.5em minus 0.4em\relax IEEE, 2022, pp. 6115--6120.

\bibitem{10579124}
V.~Murgai, V.~R.~R. Lolabhattu, R.~Stimpson, E.~Tripathi, and S.~Chickala, ``Securing the metaverse: Traffic application classification and anomaly detection,'' in \emph{2024 IEEE 25th International Symposium on a World of Wireless, Mobile and Multimedia Networks (WoWMoM)}, 2024, pp. 111--117.

\bibitem{data_1}
\BIBentryALTinterwordspacing
Y.~S. Kuruba~Manjunath, L.~Zhao, and X.-P. Zhang, ``Metaverse network traffic for classification and prediction,'' 2024. [Online]. Available: \url{https://dx.doi.org/10.21227/0qs9-f852}
\BIBentrySTDinterwordspacing

\bibitem{yogasuha5}
Y.~S.~K. Manjunath and A.~Wissborn, ``Discern-xr: An online classifier for metaverse network traffic classifcation,'' \url{https://github.com/yoga-suhas-km/Discern-XR}, (Accessed on 10/11/2024).

\bibitem{MetaQues31}
Meta, ``Meta quest 2,'' \url{https://www.meta.com/ca/quest/products/quest-2/}, (Accessed on 09/29/2024).

\bibitem{HomeVirt30}
V.~Desktop, ``Virtual desktop streamer,'' \url{https://www.vrdesktop.net/}, (Accessed on 09/29/2024).

\bibitem{wireshark}
A.~Orebaugh, G.~Ramirez, J.~Beale, and J.~Wright, \emph{Wireshark \& Ethereal Network Protocol Analyzer Toolkit}.\hskip 1em plus 0.5em minus 0.4em\relax Syngress Publishing, 2007.

\bibitem{9685808}
S.~Zhao, H.~Abou-zeid, R.~Atawia, Y.~S.~K. Manjunath, A.~B. Sediq, and X.-P. Zhang, ``Virtual reality gaming on the cloud: A reality check,'' in \emph{2021 IEEE Global Communications Conference (GLOBECOM)}, 2021, pp. 1--6.

\bibitem{biau2016random}
G.~Biau and E.~Scornet, ``A random forest guided tour,'' \emph{Test}, vol.~25, pp. 197--227, 2016.

\bibitem{questset}
\BIBentryALTinterwordspacing
S.~Baldoni, F.~Battisti, F.~Chiariotti, F.~Mistrorigo, A.~B. Shofi, P.~Testolina, A.~Traspadini, A.~Zanella, and M.~Zorzi, ``Questset: A vr dataset for network and quality of experience studies,'' in \emph{Proceedings of the 15th ACM Multimedia Systems Conference}, ser. MMSys '24, 2024, p. 408–414. [Online]. Available: \url{https://doi.org/10.1145/3625468.3652187}
\BIBentrySTDinterwordspacing

\end{thebibliography}

\end{document}